\documentclass{article}

\usepackage[preprint]{neurips_2025}
\usepackage{svg}
 \usepackage{multirow} 

\usepackage[utf8]{inputenc} 
\usepackage[T1]{fontenc}    
\usepackage{hyperref}       
\usepackage{url}            
\usepackage{booktabs}       
\usepackage{amsfonts}       
\usepackage{nicefrac}       
\usepackage{microtype}      
\usepackage{xcolor}         
\usepackage{amsmath}
\usepackage{amsmath,amssymb,amsthm}
\usepackage{graphicx} 
\usepackage{algorithm}
\usepackage{algpseudocode}
\usepackage{amsmath}
\usepackage{float}

\usepackage{tikz}
\usepackage{amsmath}
\usepackage{xcolor}
\usetikzlibrary{shapes.geometric, arrows, positioning, fit, backgrounds, calc, decorations.pathreplacing, matrix}

\newcommand{\wersa}{\textsc{WERSA}}
\usepackage{booktabs, multirow, tabularx, siunitx}

\newcolumntype{Y}{>{\raggedleft\arraybackslash}X} 

\title{Scaling Attention to Very Long Sequences in Linear Time with Wavelet-Enhanced Random Spectral Attention (\wersa)}

\author{%
  Vincenzo Dentamaro \\
  Department of Computer Science\\
  University of Bari Aldo Moro\\
  Bari, IT 70125 \\
  \texttt{vincenzo.dentamaro@uniba.it} \\
}

\begin{document}

\maketitle

\begin{abstract}
Transformer models are computationally costly on long sequences since regular attention has quadratic $O(n^2)$ time complexity. We introduce Wavelet-Enhanced Random Spectral Attention (\wersa), a novel mechanism of linear $O(n)$ time complexity that is pivotal to enable successful long-sequence processing without the performance trade-off. \wersa\ merges content-adaptive random spectral features together with multi-resolution Haar wavelets and learnable parameters to selectively attend to informative scales of data while preserving linear efficiency.

Large-scale comparisons \textbf{on single GPU} and across various benchmarks (vision, NLP, hierarchical reasoning) and various attention mechanisms (like Multiheaded Attention, Flash-Attention-2, FNet, Linformer, Performer, Waveformer), reveal uniform advantages of \wersa. It achieves best accuracy in all tests. On ArXiv classification, \wersa\ improves accuracy over vanilla attention by 1.2\% (86.2\% vs 85.0\%) while cutting training time by 81\% (296s vs 1554s) and FLOPS by 73.4\% (26.2G vs 98.4G). Significantly, \wersa\ excels where vanilla and FlashAttention-2 fail: on ArXiv-128k's extremely lengthy sequences, it achieves best accuracy (79.1\%) and AUC (0.979) among viable methods, operating on data that gives Out-Of-Memory errors to quadratic methods while being \textbf{twice as fast} as Waveformer, its next-best competitor.

By significantly reducing computational loads without compromising accuracy, \wersa\ makes possible more practical, more affordable, long-context models, in particular on low-resource hardware, for more sustainable and more scalable AI development.

\end{abstract}

\section{Introduction}
Transformer models \citep{vaswani2017attention} have drastically changed machine learning field by providing better performance in a huge variety of applications, ranging from natural language processing \citep{devlin2019bert, brown2020language} to computer vision \citep{dosovitskiy2020image, liu2021swin} and multimodal learning \citep{radford2021learning, alayrac2022flamingo}. Underlying such success is the mechanism of attention, which allows long-range dependencies to be captured efficiently by calculating interactions between each pair of tokens in a sequence. The vanilla mechanism of self-attention, however, calculates a quadratically increasing
computational cost in terms of sequence length, thereby limiting transformer applications in very long sequence environments.
This computational burden has spurred a lot of work in designing effective attention mechanisms that capture the expressiveness of regular attention with much fewer computational resources. Recent techniques include sparse attention patterns \citep{beltagy2020longformer, zaheer2020big}, low-rank approximations \citep{wang2020linformer, choromanski2020rethinking}, and kernel-based methods \citep{katharopoulos2020transformers, peng2021random}
\citep{lee2021fnet, gu2022efficiently}. Transformers have also been constructed with Fast Fourier Transform (FFT) to calculate attention in the frequency domain to minimize computational complexity from $\mathcal{O}(n^2)$ to $\mathcal{O}(n\log n)$. Such techniques do, however, rely on global sinusoidal bases, which do not capture the interaction between local and global patterns \citep{lee2021fnet}.

This work presents Wavelet-Enhanced Random Spectral Attention (\wersa), a novel attention mechanism incorporating wavelet transforms with random feature mappings to achieve a linear computational complexity of $\mathcal{O}(n)$ without compromising representational power. Unlike FFT-based methods using static sinusoidal bases, \wersa\ uses the Haar wavelet transform to provide multi-resolution capabilities, thus supporting adaptive learning of both local and global interactions. Additionally, it uses content-adaptive filtering, where input-dependent coefficients control wavelet scales through learnable parameters, enabling the model to focus on relevant frequency components.

\wersa's theoretical foundations include multi-resolution analysis allowing fine-grained attentional processing across scales, with proven exponentially decreasing approximation error as wavelet levels increase. The random feature approximation ensures $\mathcal{O}(n)$ complexity with theoretical guarantees, while trainable bandwidth parameters optimize kernel approximation. 

The scale-specific processing is particularly useful for applications where signals contain important information at different frequencies, thus improving the model's ability to capture detailed nuances and overall trends simultaneously.

The major contributions are:
\begin{enumerate}
\item \wersa: A novel $\mathcal{O}(n)$ attention mechanism maintaining performance comparable to standard quadratic attention
\item The application of multi-scale wavelet processing successfully captures both local and global dependencies.
\item Context-adaptive filtering reduces noise by applying input-dependent gating functions.
\item Empirical validation showing 73\% FLOPS reduction and 81\% faster computation while maintaining accuracy
\item Enabling the theoretical processing of very long sequences which are impractical with traditional attention mechanisms 

\end{enumerate}

\wersa\ reduces the computational cost that comes with long sequence inputs, thus reducing energy consumption and promoting the next generation AI sustainable models.

The paper is thought to be self-contained, i.e. all information required to understand the logic of \wersa\ are described here. Thus this work is organized as follows: a state-of-the-art review on efficient attention mechanisms for transformers is provided in Section 2. The WERSA method and the \wersa\ approximation theorem are presented in Section 3. The theorem proves asymptotic convergence to standard multi-headed attention but tackling very long sequences (millions of tokens of context) in linear time. Datasets used and experimental setups are shown in Section 4. Results and their discussion are provided in Section 5, while conclusions are provided in Section 6.

Additionally, background information of attention mechanisms and Wavelet Transform are provided in \textbf{Appendix A}.  \textbf{Appendix B} contains the proof of \wersa\ approximation theorem. \textbf{Appendix C } shows Haar Wavelet transform for generalized levels. The \wersa\ pseudo-algorithm is sketched in \textbf{Appendix D} while ablation studies are shown in \textbf{Appendix E}.

\section{Literature Review}

The quadratic complexity of standard multi-headed attention (MHA) of the base transformer model has opened various ways to enhance it. \citet{tay2022efficient} provide a comprehensive overview of efficient transformer models and classify them into various different strategies.

Sparse Attention techniques minimize computational overhead by constraining attention mechanisms to a specific subset of positions. Some notable examples include Sparse Transformer \citep{child2019generating}, Longformer \citep{beltagy2020longformer}, and BigBird \citep{zaheer2020big}. These approaches typically achieve $\mathcal{O}(n \sqrt{n})$ or $\mathcal{O}(n \log n)$ complexity.

Low-Rank Approximations utilize low-rank factorizations to approximate the attention matrix. The Linformer \citep{wang2020linformer} truncates the length dimension to a fixed size, which results in $\mathcal{O}(n)$ complexity. Performer \citep{choromanski2020rethinking} and Linear Transformer \citep{katharopoulos2020transformers} utilize kernel approximations to approximate softmax attention and achieve linear complexity by avoiding the explicit computation of the quadratic-sized attention matrix. 

Recurrence-based Methods utilize recurrent models to efficiently capture long-term dependencies. Segment recurrence has been utilized by Transformer-XL \citep{dai2019transformer} and Compressive Transformer \citep{rae2019compressive}, while Linear Recurrent Units \citep{orvieto2023resurrecting} and RWKV \citep{peng2023rwkv} use a combination of recurrent processing and parallel computation for efficient inference.

\citet{dao2022flashattention} presented FlashAttention, a method to reduce memory access patterns to enable exact attention with lower memory bandwidth overhead. FlashAttention-2 \citep{dao2023flashattention}, a variant, supported successful training and inference with long-length inputs. These methods, with asymptotic complexity being $\mathcal{O}(n^2)$, have a much lowered real-world computational cost. \citet{xiong2021nystromformer} introduced Nyström attention with computational complexity of  $\mathcal{O}(n \sqrt{n})$. MEGA \citep{ma2022mega}, however, blended exponential moving average with gated attention and attained $\mathcal{O}(n)$ complexity with long-term learning performance in language modeling tasks.
 
Fourier-based Methods leverage the effectiveness of Fast Fourier Transform. FNet \citep{lee2021fnet} substitutes attention computation with Fourier transforms where attention is computed in the frequency space with a time complexity of $\mathcal{O}(n \log n)$. \citet{gu2022efficiently} provide a mechanism using FFT in the frequency domain for attention. Another recent method by \citet{zhou2022fedformer} provides a hybrid method using a combination of FFT and low-rank approximations in the frequency domain to enhance transformers for long inputs.
 
Frequency-domain techniques include  \citet{li2023fourier} who proposed Fourier Neural Operators (FNO) to perform computations in spectral space, but also the Waveformer \citep{zhuang2022waveformer} which is a transformer that applies discrete wavelet transformation (DWT) on multi-resolution sequences. It decomposes input into approximated coefficients  through Haar wavelet transformation similarly to the more recent WavSpa \citep{zhuang2024wavspa}.

The proposed \wersa\ offers important theoretical and architectural advances over Waveformer \citep{zhuang2022waveformer} and WavSpa \citep{zhuang2024wavspa}: (1) it employs content-adaptive filtering, with wavelet coefficients dynamically computed from input features, as opposed to fixed transforms in Waveformer; (2) it employs learnable, scale-dependent parameters for adaptive multi-resolution analysis as opposed to WavSpa \citep{zhuang2024wavspa}; (3) it achieves strict \( \mathcal{O}(n) \) complexity via random feature projections and multi-scale decomposition, as opposed to Waveformer’s quadratic frequency cost; and (4) it offers formal, logarithmically scaling error bounds for attention approximation. These advances enhance \wersa’s capacity to model efficiently both local and global dependencies over long sequences.

\section{Wavelet-Enhanced Random Spectral Attention}

\wersa\ combines a set of central principles:

\begin{enumerate}
\item \textbf{Multi-Resolution Analysis}: Employing wavelet transforms to partition input into various scales, and thereby maintaining local and global relationships.
\item \textbf{Sparse Representation}: Taking advantage of natural sparsity in wavelet representation, particularly in signals with local structure.
\item \textbf{Adaptive Filtering}: Weighing particular wavelet coefficients in terms of input content.
\item \textbf{Random Feature Approximation}: Linearization of the attention computation by random feature projection for complexity reduction.
\item \textbf{Multi-Scale Fusion}: Processing wavelet scales with efficient filtering and combining them with learnable gates to enhance representation power.
\item \textbf{Unified Scale Attention}: Computing attention with scale-filtered signals for better computational efficiency.
\item \textbf{Bandwidth Adaptation}: Using trainable bandwidth parameters to optimize the kernel approximation for the data distribution.
\end{enumerate}
\subsection{\wersa\ Algorithm}
\begin{figure}[H]
\centering
\begin{tikzpicture}[
    block/.style={rectangle, draw, fill=blue!20, 
        text width=1.8cm, text centered, rounded corners, minimum height=1cm, font=\scriptsize},
    line/.style={draw, -latex', thin},
    wavelet/.style={rectangle, draw, fill=orange!30, 
        text width=1.8cm, text centered, rounded corners, minimum height=1cm, font=\scriptsize},
    attention/.style={rectangle, draw, fill=green!20, 
        text width=1.9cm, text centered, rounded corners, minimum height=1cm, font=\scriptsize},
    data/.style={rectangle, draw, fill=gray!10, 
        text width=1.5cm, text centered, minimum height=0.8cm, font=\scriptsize},
    scale=0.7, every node/.style={scale=0.7}
]

\begin{scope}
    \node[data] (input) at (0,0) {$Q, K, V \in$$\mathbb{R}^{n \times d}$};

    \node[block, right=0.2cm of input] (projection) {Linear\\Projection$Q' = QW^Q$$K' = KW^K$$V' = VW^V$};

    \node[block, right=0.2cm of projection] (splitting) {Multi-Head Splitting\\Reshape to $\mathbb{R}^{n \times h \times d_h}$};

    \begin{pgfonlayer}{background}
        \node[rectangle, draw, thick, rounded corners, fill=violet!10, 
              fit={(input) (projection) (splitting)}, inner sep=0.12cm] {};
    \end{pgfonlayer}
\end{scope}

\begin{scope}
    \node[wavelet, right=0.2cm of splitting] (wavelet) {Wavelet\\Transform$Q^{WT} = \mathcal{W}(Q')$$K^{WT} = \mathcal{W}(K')$};

    \node[wavelet, right=0.2cm of wavelet] (filtering) {Adaptive\\Filtering$F_i =$$\sigma(g(Q'_{avg}))$};

    \node[wavelet, right=0.2cm of filtering] (reconstruction) {Reconstruction$Q_F =$$\mathcal{W}^{-1}(F \odot$$Q^{WT})$};

    \begin{pgfonlayer}{background}
        \node[rectangle, draw, thick, rounded corners, fill=orange!10, 
              fit={(wavelet) (filtering) (reconstruction)}, inner sep=0.12cm] {};
    \end{pgfonlayer}
\end{scope}

\begin{scope}
    \node[attention, right=0.5cm of reconstruction] (random) {Random\\Feature$\phi(x) =$$\text{ReLU}(xR/\beta)$};

    \node[attention, below=0.2cm of random] (att_comp) {Attention$\phi(Q_F)$$(\phi(K_F)^TV')$};

    \node[attention, right=0.2cm of random] (multi_head) {Multi-Head\\Combine};

    \node[data, right=0.2cm of multi_head] (output) {Output};

    \begin{pgfonlayer}{background}
        \node[rectangle, draw, thick, rounded corners, fill=green!10, 
              fit={(random) (att_comp) (multi_head)}, inner sep=0.12cm] {};
    \end{pgfonlayer}
\end{scope}

\path[line] (input) -- (projection);
\path[line] (projection) -- (splitting);
\path[line] (splitting) -- (wavelet);
\path[line] (wavelet) -- (filtering);
\path[line] (filtering) -- (reconstruction);

\coordinate (branch_point) at ($(reconstruction.east)!0.5!(random.west)$);
\path[line] (reconstruction) -- (random);
\path[line] (reconstruction.east) -| ($(random.west) + (-0.3,-0.6)$) -- (att_comp.west);

\path[line] (random) -- (multi_head);
\path[line] (att_comp) -- ($(att_comp.east)+(0.2,0)$) |- (multi_head);

\path[line] (multi_head) -- (output);

\end{tikzpicture}
\caption{Wavelet Enhanced Random Feature Self-Attention (\wersa) architecture. From left to right: Input Processing (gray/blue), Wavelet Processing (orange), and Attention Mechanism (green).}
\label{fig:wersa}
\end{figure}
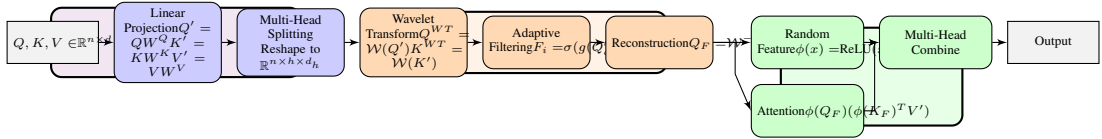

Let queries $Q \in \mathbb{R}^{n \times d}$, keys $K \in \mathbb{R}^{n \times d}$, and values $V \in \mathbb{R}^{n \times d}$, the \wersa\ mechanism is as follows:
\begin{enumerate}
   \item \textbf{Linear Projection:} Computes projected queries, keys, and values: 
$$
Q' = QW^Q,\quad K' = KW^K,\quad V' = VW^V,
$$
where $W^Q, W^K, W^V \in \mathbb{R}^{d \times h \cdot d_k}$.
    
    \item \textbf{Multi-Head Splitting:} Divides $Q'$, $K'$, and $V'$ into $h$ heads and reshape them to $\mathbb{R}^{n \times h \times d_h}$, where $d_h = d_k/h$.
    
    \item \textbf{Wavelet Decomposition:} Performs the wavelet transform (for simple implementation, it has been used the Haar wavelet transform see Appendix A) on $Q'$ and $K'$ for $L$ levels:
    \[
    Q^{WT} = \mathcal{W}(Q'), \quad K^{WT} = \mathcal{W}(K'),
    \]
    where $\mathcal{W}$ generates a sequence of wavelet coefficients at various scales.
    It is important to state that the method is generic and thus Haar can be substituted with other types of wavelet decomposition such as Daubechies, symlets, etc. 
    
    \item \textbf{Adaptive Wavelet Filtering:} One of \wersa's innovations includes adaptive wavelet filtering. The average query representation for a specific head $h$ can be computed as:
\begin{equation}
Q'_{\text{avg}, h} = \frac{1}{n} \sum_{i=1}^{n} Q'_{h, i}.
\end{equation}
A neural network represented by $g$ then maps $Q'_{\text{avg}, h}$ to filter coefficients:
\begin{equation}
F = \sigma(g(Q'_{\text{avg}, h})),
\end{equation}
where $\sigma$ denotes the sigmoid function,yielding coefficients in the $[0,1]$ range. The coefficients are then applied element-wise to each respective wavelet coefficient acting like a filter. Additionally, each wavelet scale is modulated by a learnable parameter:
\begin{equation}
F_i = \sigma\bigl(g(Q'_{\text{avg}, h})_i\bigr) \cdot \omega_i,
\end{equation}
where $\omega_i$ controls the importance of the $i$-th wavelet scale.
Prior wavelet-based transformers \citep{zhuang2022waveformer} process all wavelet scales equally, which is suboptimal for sequences with heterogeneous frequency components. \wersa's gating mechanism $\omega_i$ learns to suppress noisy high-frequency scales or enhance low-frequency global patterns, depending on the input.

    \item \textbf{Filtered Wavelet Representation:} Multiplies the wavelet coefficients by the corresponding filter coefficients:
   \[
F_i \;=\; \sigma\!\Bigl(g\bigl(Q'_{\text{avg}, h}\bigr)_i\Bigr) \;\times\; \omega_i,
\]
    
    \item \textbf{Unified Multi-Scale Reconstruction:} \wersa\ efficiently reconstructs filtered signals through a single operation:
    \[
    Q_F = \mathcal{W}^{-1}(F \odot Q^{WT}), \quad K_F = \mathcal{W}^{-1}(F \odot K^{WT}),
    \]
    where $\mathcal{W}^{-1}$ represents the inverse wavelet transform of the filtered coefficients.
    
    \item \textbf{Random Feature Projection:} To mitigate computational complexity, approximations with random features are utilized to approximate the softmax kernel. This is done by using a map $\phi(x)$:
\begin{equation}
K(x,y) \approx \phi(x)^T\phi(y),
\end{equation}
and specifically, a ReLU-based random feature map is used:
\begin{equation}
\phi(x) = \text{ReLU}(xR),
\end{equation}
with $R$ randomly drawn from $\mathbb{R}^{d_h \times m}$. The projection is enhanced with a trainable bandwidth parameter $\beta$:
\begin{equation}
\phi(x) = \text{ReLU}(xR / \beta),
\end{equation}
where $R$ can be initialized as orthogonal or Gaussian matrices. This yields the linear attention approximation:
\begin{equation}
\text{softmax}(QK^T)V \approx \frac{\phi(Q_F)(\phi(K_F)^TV)}{\phi(Q_F)\phi(K_F)^T\mathbf{1} + \epsilon}.
\end{equation}
    
    \item \textbf{Attention Computation:} Compute attention with the filtered signals:
    \[
    \text{Attention} = \frac{\phi(Q_F)(\phi(K_F)^TV')}{\phi(Q_F)\phi(K_F)^T\mathbf{1} + \epsilon},
    \]
    where $\epsilon$ is a small constant aiming at ensuring numerical stability.
    
    \item \textbf{Multi-Head Combination:} Concatenates the outputs of all heads and project them:
    \[
    \text{\wersa}(Q, K, V) = \text{Concat}(\text{Attention}_1, \ldots, \text{Attention}_h)W^\textit{Output} ,
    \]
    where $W^\textit{Output} \in \mathbb{R}^{d_k \times d}$.
\end{enumerate}
A pictorial representation of the operations flow is depicted in Figure \ref{fig:wersa}.

In order to increase transparency and reproducibility the \wersa\ pseudocode can be found in \textbf{Appendix D} while the Hugging Face Transformers compatible implementation will be released upon acceptance.

\subsection{Computational Complexity Analysis}
The computational complexity of \wersa\ is derived by analysing each component:
\begin{enumerate}
    \item \textbf{Linear Projection}: $\mathcal{O}(nd^2)$.
    \item \textbf{Wavelet Decomposition and Reconstruction}: $\mathcal{O}(nd)$ because vectorized Haar transform has a linear time complexity.
    \item \textbf{Adaptive Filtering}: $\mathcal{O}(d^2)$ for filter computation and $\mathcal{O}(nd)$ for filtering.
    \item \textbf{Random Feature Projection}: $\mathcal{O}(ndm)$ with fixed dimension $m$.
    \item \textbf{Linear Attention Computation}: $\mathcal{O}(ndm)$.
    \item \textbf{Final Projection}: $\mathcal{O}(nd^2)$.
    \item \textbf{Unified Filtered Processing}: $\mathcal{O}(nd)$ for reconstruction with filtering, reducing the multiplier of $L$ from the original approach.
\end{enumerate}
Thus, the overall aggregated complexity is 
\[
\mathcal{O}(nd^2 + nd + d^2 + ndm + ndm + nd^2 + nd) = \mathcal{O}(nd(d+m)),
\]
which, since $d$ and $m$ are constants with respect to $n$, this reduces to $\mathcal{O}(n)$. The unified filtering approach maintains the overall $\mathcal{O}(n)$ complexity while reducing the computational burden compared to separate scale processing.

In practice, it can be observed from results reported in Table \ref{tab:unified_results} that for small-scale models or very short sequences, the constants used by wavelet decomposition and random feature projections can override the asymptotic gain, leading to slightly slower training.

\subsection{\wersa\ Long-Context Approximation Theorem}

Since computational resources are generally a limiting factor regarding the scaling of attention mechanisms for very long sequences, it is possible to estabilish the theoretical foundation of \wersa\ for tackling very long sequences formally, through an approximation theorem.

\wersa\ enhances multi-head attention (MHA) for long sequences using \textbf{multi-scale hierarchical processing} with wavelet transforms to divide sequences into localized  and global patterns, maintaining complex details without requiring quadratic token comparisons. In addition, \textbf{context-adaptive filtering} removes noise and reinforces important interactions using input-based gating and randomized projections, resulting in complexity of $\mathcal{O}(n)$. While MHA becomes \textit{computationally infeasible} for several million-token sequences \citep{fan2024rmt}, \wersa\ maintains accuracy while providing significantly improved efficiency, enabling scalable long-context modeling. The following theorem shows these properties.

\textbf{\wersa\ Long-Context Approximation Theorem}

Let $ Q, K, V \in \mathbb{R}^{n \times d} $ be the query, key, and value matrices, respectively. It is possible to assume that \wersa\ uses $ L $-level wavelet decomposition (in the experiments $L$ is fixed to 2), a fixed random projection dimension $ m $, and a fixed bandwidth parameter $ \beta $. In the experiments $m$ is 1024 and $\beta$ is 1.0.
\begin{enumerate}
    \item \textbf{Approximation Guarantee:} For all $ \epsilon > 0 $ and for any given $ \delta \in (0,1) $, if 
    $$
    m \geq \frac{16}{\epsilon^2} \log\left(\frac{2n^2}{\delta}\right),
    $$
    Where $16$ constant comes from McDiarmid’s Inequality computation to balance theoretical safety and efficiency avoiding having randomized projection with a very high number of features and useless overhead.
    Hence, with at least a probability of $1 - \delta$,
    $$
    \left\| \mathrm{\wersa}(Q, K, V) - \mathrm{Attention}(Q, K, V) \right\|_F \leq \epsilon \|V\|_F + \mathcal{O}(2^{-\alpha L}),
    $$
    The value of $ \alpha > 0 $ also depends on the smooth conditions for $ Q $ and $ K $, i.e. $ \alpha$ decay rate increases when $Q$ and $K$ are less noisy. The former ($\epsilon \|V\|_F$) is due to random feature representation, and the latter ($\mathcal{O}(2^{-\alpha L})$) is due to wavelet truncation.
    
    \item \textbf{Linear Complexity:} For given constants $ d, m, L $, \wersa's computational complexity is 
    $$
    \mathcal{O}(n d(d + m + L)),
    $$
    This reduces to $ \mathcal{O}(n) $ if $ d, m, L $ are treated as constant multiples (i.e. if the Wavelet level $L$ is fixed as $d,m$ are fixed parameters). For $ L = \mathcal{O}(\log n) $, the complexity is $ \mathcal{O}(n \log n) $.
    
    \item \textbf{Optimality of Adaptive Filters:} Define $ \mathcal{F}^* $ as the optimal wavelet filter with respect to minimizing $ \|A(Q,K) - \hat{A}(Q_F,K_F)\|_F $. Then, the obtained filter $ \hat{F} $ satisfies:
    $$
    \| \hat{F} - \mathcal{F}^* \|_F \leq \mathcal{O}\left(\sqrt{\frac{d \log n}{n}}\right),
    $$
    to produce a limiting value as a function approaches infinity.
\end{enumerate} 
This theorem states that \wersa\ approximates regular attention with bounded error, attains linear time complexity, and guarantees near-optimal filter selection, rendering it theoretically sound and computationally efficient to process long sequences. 

The proof is provided in \textbf{Appendix B}.

\section{Datasets and Experimental Setups}
One of the goals of this paper was to create an attention mechanism capable of being trained on single video card. All the experiments were executed on a PC with AMD Ryzen™ Threadripper™ 3970X × 64 with 128GB of DDR4 RAM, \textbf{a single Nvidia A6000 GPU with 48GB DDR6 VRAM} and with Ubuntu 24.04.2 LTS. The experimental comparison involves the use of transformer architectures with few architectural changes depending on the problem. For each dataset the exact same transformer architecture was used, changing only the attention mechanism type. Seven types of attention mechanisms are benchmarked both in terms of accuracy performances and speed: multi-headed attention (MHA) \citet{vaswani2017attention}, FNet \citet{lee2021fnet}, FlashAttention-2 \citep{dao2023flashattention}, Linformer \citep{wang2020linformer}, Performer \citep{choromanski2020rethinking}, Waveformer \citep{zhuang2022waveformer} and the novel Wavelet-Enhanced Random Spectral Attention (\wersa). All experiments were performed 5 times and performances average as well as standard deviations are reported.

\subsection{IMDB Movie Review Dataset}

IMDB dataset \citep{maas2011learning} consists of 50,000 polarized movie reviews with scores $\leq 4$ or $\geq 7$ out of a total of 10 with balanced test/training split. Preprocessing included capping to top 20,000 frequent words, standardization to 4,000 tokens, 25,000 instances have been used for training and 25,000 for testing. Validation has been set to 25\% of training set.
For this sentiment classification task, a vanilla and simple transformer architecture was implemented, consisting of 2 encoder layers with a 128-dimensional embedding space, 4 attention heads, and feed-forward networks with an internal dimension of 256, trained over 100 epochs with batch size 32.

\subsection{CIFAR 10 \& 100}
The CIFAR-100 dataset \citet{krizhevsky2009learning} is composed of 60,000 color images of size 32×32 and contains 100 fine-grained classes. CIFAR-10 is of equal number of images (60,000) distributed across 10 classes with each class containing 6,000 images. Both are difficult owing to low spatial resolution though CIFAR-100 is more difficult with higher intra-class variation and inter-class similarity. 80\% of each of these train set was used for training, 20\% of training data was used for validation and the resulting 20\% for test.

To evaluate \wersa\ in the context of computer vision, the same Vision Transformer (ViT) architecture was implemented for both CIFAR-10 and CIFAR-100 as in \citep{dosovitskiy2020image}. The framework divides each image into non-overlapping patches of size $p \times p \times 3$ (in this case $p$ is 4), resulting in a sequence of $(32/p)^2$ elements. These patches are embedded into $d$-dimensional space, with positional encodings added, and then processed by transformer encoder layers.

All models utilized identical architecture except for the attention mechanism, with Adam optimizer (learning rate 1e-4), batch size 16, early stopping, $d_{\text{model}}=128$, 4 attention heads, and 2 encoder layers.

\subsection{ListOps Dataset}

The ListOps dataset \citep{nangia2018listops} tests models on hierarchically structured inputs as nested math operations (e.g., \texttt{[MAX 2 9 [MIN 4 7] 0])}. The task is to test a model’s ability to parse hierarchical and long-range dependencies. Preprocess was done by tokenizing expressions, encoding tokens by integer IDs and padding to fixed length. Sequences were then represented as classification problem across possible output values. Transformer architecture employed a 4-layer encoder architecture with an embedding dimension of 256, 8 parallel attention heads, and feed-forward networks with an internal dimension of 1024. Training was done with Adam optimizer with batch size of 4 for 70 epochs. Official train, validation and test splits were used for the benchmark, i.e. 90000 examples in the training set, 10800 in validation set (20\% of training set) and 10000 in the test sets.

\subsection{ArXiv Dataset} 

The ArXiv dataset \citep{clement2019use} is comprised of scientific paper abstracts labeled in multiple domains and is a challenging text classification problem. Pre-processing was done with text cleaning and tokenization using 20k subword vocabulary method and shaping to fixed-sized sequences using right padding and clipping. The same transformer architecture used for ListOps was used for this problem without modifying configuration parameters. Such consistency in architecture enables comparison of model capability on processing for sequences. Original split have been used: 33000 Arxiv Papers divided into 3 splits: train (28,000), val (2,500) and test (2,500).  

\subsection{ArXiv-128k Dataset} 

The dataset is exactly the same as the ArXiv dataset presented previously, but it has been tokenized with a length of 128k tokens. The encoder only Transformer architecture employed is similar to the one used for ListOps but this time, due to computational constraints, it had just 2 layers, 64 embedding dimensions,  4 parallel heads and the feed-forward network had only 256 dimensions. Training was conducted with batch size of two and for just 25 epochs.

\section{Results and Discussion}
\begin{table}[h] 
\caption{Mean $\pm$ standard deviation over five seeds.}
\label{tab:unified_results}  
\centering 
\resizebox{\textwidth}{!}{
\begin{tabular}{lccccccrr}
\toprule
\textbf{Dataset} & \textbf{Att Type} &
\textbf{Acc} & \textbf{Prec} & \textbf{Rec} & \textbf{F1} & \textbf{AUC} &
\textbf{Time (s)} & \textbf{FLOPS (G)} \\
\midrule
\multirow{7}{*}{\scriptsize CIFAR-100}
& ViT-Standard & 0.3804$\pm$0.0035 & 0.3748$\pm$0.0032 & 0.3804$\pm$0.0033 & 0.3735$\pm$0.0034 & 0.9450$\pm$0.0010 & 8.00$\pm$0.08 & 0.0356 \\
& ViT-FlashAttn-2 & 0.3695$\pm$0.0031 & 0.3657$\pm$0.0030 & 0.3695$\pm$0.0031 & 0.3608$\pm$0.0029 & 0.9408$\pm$0.0011 & \textbf{6.15}$\pm$0.05 & 0.0354 \\
& ViT-FNet & 0.3234$\pm$0.0028 & 0.3108$\pm$0.0026 & 0.3234$\pm$0.0027 & 0.3091$\pm$0.0026 & 0.9133$\pm$0.0014 & 7.00$\pm$0.07 & 0.0323 \\
& ViT-Linform & 0.3590$\pm$0.0032 & 0.3481$\pm$0.0031 & 0.3590$\pm$0.0032 & 0.3497$\pm$0.0031 & 0.9364$\pm$0.0012 & 7.00$\pm$0.06 & \textbf{0.0321} \\
& ViT-Perform & 0.3558$\pm$0.0030 & 0.3457$\pm$0.0028 & 0.3558$\pm$0.0031 & 0.3462$\pm$0.0029 & 0.9357$\pm$0.0013 & 7.26$\pm$0.07 & 0.0323 \\
& ViT-Waveform & 0.3893$\pm$0.0036 & 0.3794$\pm$0.0035 & 0.3893$\pm$0.0035 & 0.3792$\pm$0.0034 & 0.9483$\pm$0.0010 & 11.00$\pm$0.11 & 0.0374 \\
& ViT-\textbf{WERSA} & \textbf{0.3901}$\pm$0.0034 & \textbf{0.3801}$\pm$0.0033 & \textbf{0.3901}$\pm$0.0034 & \textbf{0.3800}$\pm$0.0033 & \textbf{0.9485}$\pm$0.0009 & 10.62$\pm$0.10 & 0.0365 \\
\midrule
\multirow{7}{*}{\scriptsize CIFAR-10}
& ViT-Standard & 0.8198$\pm$0.0020 & 0.8174$\pm$0.0021 & 0.8198$\pm$0.0020 & 0.8185$\pm$0.0021 & 0.9711$\pm$0.0007 & 8.00$\pm$0.08 & 0.0356 \\
& ViT-FlashAttn-2 & 0.8160$\pm$0.0021 & 0.8144$\pm$0.0020 & 0.8160$\pm$0.0021 & 0.8148$\pm$0.0020 & 0.9685$\pm$0.0008 & 6.15$\pm$0.05 & 0.0354 \\
& ViT-FNet & 0.7264$\pm$0.0025 & 0.7170$\pm$0.0024 & 0.7264$\pm$0.0025 & 0.7199$\pm$0.0024 & 0.9275$\pm$0.0012 & \textbf{2.00}$\pm$0.03 & 0.0323 \\
& ViT-Linform & 0.8122$\pm$0.0020 & 0.8095$\pm$0.0020 & 0.8122$\pm$0.0021 & 0.8110$\pm$0.0020 & 0.9673$\pm$0.0008 & 7.00$\pm$0.06 & 0.0321 \\
& ViT-Perform & 0.8139$\pm$0.0021 & 0.8113$\pm$0.0021 & 0.8139$\pm$0.0020 & 0.8127$\pm$0.0021 & 0.9680$\pm$0.0008 & 7.26$\pm$0.07 & 0.0323 \\
& ViT-Waveform & 0.8178$\pm$0.0022 & 0.8154$\pm$0.0021 & 0.8178$\pm$0.0022 & 0.8165$\pm$0.0021 & 0.9690$\pm$0.0008 & 3.22$\pm$0.04 & \textbf{0.0306} \\
& ViT-\textbf{WERSA} & \textbf{0.8298}$\pm$0.0020 & \textbf{0.8267}$\pm$0.0020 & \textbf{0.8298}$\pm$0.0020 & \textbf{0.8285}$\pm$0.0020 & \textbf{0.9737}$\pm$0.0007 & 3.50$\pm$0.05 & 0.0322 \\
\midrule
\multirow{7}{*}{\scriptsize IMDB}
& Standard & 0.8892$\pm$0.0017 & 0.8894$\pm$0.0017 & 0.8892$\pm$0.0018 & 0.8892$\pm$0.0017 & 0.9561$\pm$0.0006 & 363.0$\pm$3.4 & 9.18 \\
& FlashAttn-2 & 0.8861$\pm$0.0018 & 0.8861$\pm$0.0018 & 0.8861$\pm$0.0018 & 0.8861$\pm$0.0018 & 0.9548$\pm$0.0007 & 279.2$\pm$2.6 & 9.12 \\
& FNet & 0.8332$\pm$0.0021 & 0.8355$\pm$0.0021 & 0.8332$\pm$0.0020 & 0.8329$\pm$0.0021 & 0.9088$\pm$0.0009 & \textbf{39.0}$\pm$0.3 & 0.543 \\
& Linformer & 0.8452$\pm$0.0020 & 0.8500$\pm$0.0019 & 0.8452$\pm$0.0021 & 0.8447$\pm$0.0020 & 0.9297$\pm$0.0008 & 60.0$\pm$0.6 & 1.09 \\
& Performer & 0.8661$\pm$0.0019 & 0.8690$\pm$0.0019 & 0.8661$\pm$0.0019 & 0.8659$\pm$0.0019 & 0.9378$\pm$0.0007 & 60.6$\pm$0.6 & 1.36 \\
& Waveformer & 0.8810$\pm$0.0018 & 0.8812$\pm$0.0017 & 0.8810$\pm$0.0018 & 0.8810$\pm$0.0018 & 0.9525$\pm$0.0007 & 39.6$\pm$0.4 & \textbf{0.534} \\
& \textbf{WERSA} & \textbf{0.8901}$\pm$0.0016 & \textbf{0.8900}$\pm$0.0016 & \textbf{0.8901}$\pm$0.0016 & \textbf{0.8901}$\pm$0.0016 & \textbf{0.9612}$\pm$0.0006 & 43.0$\pm$0.4 & 0.562 \\
\midrule
\multirow{7}{*}{\scriptsize ListOps}
& Standard & 0.4180$\pm$0.0040 & 0.4318$\pm$0.0041 & 0.4180$\pm$0.0040 & 0.3969$\pm$0.0043 & 0.8298$\pm$0.0015 & 444$\pm$4 & 98.4 \\
& FlashAttn-2 & 0.4190$\pm$0.0041 & 0.4009$\pm$0.0039 & 0.4190$\pm$0.0040 & 0.3813$\pm$0.0043 & 0.8282$\pm$0.0015 & 206$\pm$2 & 97.9 \\
& FNet & 0.2790$\pm$0.0032 & 0.0991$\pm$0.0025 & 0.2790$\pm$0.0033 & 0.1457$\pm$0.0027 & 0.6767$\pm$0.0022 & \textbf{51}$\pm$1 & 26.0 \\
& Linformer & 0.3775$\pm$0.0036 & 0.2631$\pm$0.0031 & 0.3775$\pm$0.0035 & 0.2897$\pm$0.0033 & 0.7868$\pm$0.0019 & 71.4$\pm$0.8 & 30.5 \\
& Performer & 0.4140$\pm$0.0039 & 0.3916$\pm$0.0037 & 0.4140$\pm$0.0038 & 0.3791$\pm$0.0040 & 0.8295$\pm$0.0015 & 77.2$\pm$0.9 & 32.7 \\
& Waveformer & 0.4165$\pm$0.0038 & 0.4242$\pm$0.0040 & 0.4165$\pm$0.0039 & 0.3889$\pm$0.0039 & 0.8260$\pm$0.0016 & 75.8$\pm$0.8 & \textbf{24.9} \\
& \textbf{WERSA} & \textbf{0.4233}$\pm$0.0037 & 0.4083$\pm$0.0038 & \textbf{0.4233}$\pm$0.0037 & 0.3913$\pm$0.0039 & \textbf{0.8332}$\pm$0.0015 & 82.4$\pm$0.9 & 26.2 \\
\midrule
\multirow{7}{*}{\scriptsize ArXiv}
& Standard & 0.8504$\pm$0.0025 & 0.8479$\pm$0.0025 & 0.8504$\pm$0.0026 & 0.8477$\pm$0.0025 & 0.9828$\pm$0.0004 & 1554$\pm$10 & 98.4 \\
& FlashAttn-2 & 0.8334$\pm$0.0027 & 0.8309$\pm$0.0026 & 0.8334$\pm$0.0027 & 0.8307$\pm$0.0026 & 0.9628$\pm$0.0007 & 1196$\pm$ 9 & 97.9 \\
& FNet & 0.6840$\pm$0.0032 & 0.6857$\pm$0.0031 & 0.6840$\pm$0.0032 & 0.6781$\pm$0.0034 & 0.9502$\pm$0.0009 & 346$\pm$3 & 26.0 \\
& Linformer & 0.1272$\pm$0.0020 & 0.0162$\pm$0.0012 & 0.1272$\pm$0.0020 & 0.0287$\pm$0.0015 & 0.5000$\pm$0.0000 & \textbf{252}$\pm$2 & 30.5 \\
& Performer & 0.8092$\pm$0.0029 & 0.8193$\pm$0.0029 & 0.8092$\pm$0.0029 & 0.8070$\pm$0.0029 & 0.9770$\pm$0.0005 & 316$\pm$3 & 32.7 \\
& Waveformer & 0.8488$\pm$0.0026 & 0.8507$\pm$0.0026 & 0.8488$\pm$0.0026 & 0.8481$\pm$0.0026 & 0.9849$\pm$0.0004 & 662$\pm$5 & 26.4 \\
& \textbf{WERSA} & \textbf{0.8618}$\pm$0.0024 & \textbf{0.8605}$\pm$0.0024 & \textbf{0.8618}$\pm$0.0024 & \textbf{0.8606}$\pm$0.0024 & \textbf{0.9910}$\pm$0.0003 & 296$\pm$3 & 26.2 \\
\midrule
\multirow{7}{*}{\scriptsize ArXiv-128k}
& Standard & OOM & OOM & OOM & OOM & OOM & OOM & 9330.0 \\
& FlashAttn-2 & OOM & OOM & OOM & OOM & OOM & OOM & 9200.0 \\
& FNet & 0.2016$\pm$0.0040 & 0.1908$\pm$0.0041 & 0.2016$\pm$0.0041 & 0.1610$\pm$0.0045 & 0.7660$\pm$0.0022 & \textbf{20.0}$\pm$0.3 & 26.1 \\
& Linformer & 0.0932$\pm$0.0032 & 0.0087$\pm$0.0015 & 0.0932$\pm$0.0033 & 0.0159$\pm$0.0020 & 0.4771$\pm$0.0031 & 41.0$\pm$0.5 & 44.5 \\
& Performer & 0.2996$\pm$0.0035 & 0.3146$\pm$0.0035 & 0.2996$\pm$0.0036 & 0.2706$\pm$0.0040 & 0.8348$\pm$0.0024 & 42.0$\pm$0.4 & 52.4 \\
& Waveformer & 0.7900$\pm$0.0040 & 0.7920$\pm$0.0040 & 0.7900$\pm$0.0041 & 0.7841$\pm$0.0042 & 0.9728$\pm$0.0010 & 84.0$\pm$0.9 & 28.5 \\
& \textbf{WERSA} & \textbf{0.7909}$\pm$0.0038 & \textbf{0.8151}$\pm$0.0039 & \textbf{0.7988}$\pm$0.0039 & \textbf{0.7946}$\pm$0.0040 & \textbf{0.9793}$\pm$0.0009 & 41.0$\pm$0.5 & 28.5 \\
\bottomrule
\end{tabular}%
} 
\end{table}

\renewcommand{\arraystretch}{1.0} 
The results provided in Table~\ref{tab:unified_results} illustrate \wersa's excellent tradeoff between efficiency in computations and performance metric on heterogenous datasets. \wersa\ is found to provide similar or improved accuracy compared to traditional attention mechanisms while cutting down on computational costs to a significant extent. \wersa\ attains best accuracy on CIFAR-10 dataset with 82.98\% while using merely 43.8\% of standard MHA per epoch time resources. \wersa\ shows improved accuracy by 1.14\% (86.18\% compared to 85.04\%) on arXiv dataset while showing corresponding improvement in training duration by up to 81\% (296.18s compared to 1554.43s). This tendency is consistent across datasets with \wersa\ registering best F1 scores for four out of five test datasets, and the best AUC overall.
On challenging Arxiv-128k dataset, containing extremely long sequences, \wersa\ performed best in scalability as well as performance. Regular attention as well as FlashAttention-2 both faltered by resulting in Out-Of-Memory errors, establishing evidence of the quadratic complexity boundary. \wersa, on the other hand, appropriately computed data with highest Accuracy (79.1\%), Precision (81.5\%), F1-score (79.5\%), as well as AUC (0.979) amongst all methods. FNet was faster (20s vs 41s) but a bit more FLOPS-inefficient, but its accuracy was much lower (20.2\%). \wersa\ was also much faster than Waveformer (41s vs 84s) and even a bit more accurate, establishing evidence of its significant computational cost cut and superior performance on extremely long sequences.

\wersa\ hierarchical wavelet decompositions capture multi-scale pattern; by partitioning sequential inputs into layers of different resolutions, it detects local interdependencies with high-frequency wavelets and global context with low-frequency terms. This dual ability is the key that makes \wersa\ excel on both CIFAR databases (where fine-grained grain extraction is necessary) and on document classification (where thematic understanding on a macro sense is necessary). \wersa's +17.78\% precision improvement on CIFAR-10 testifies that it excels in capturing spatial interrelationships, while its state-of-the-art accuracy on ListOps (42.33\%) testifies that it excels in capturing hierarchical dependencies that are component tasks of compositional reasoning.

On the CIFAR-100 dataset, \wersa\ technique demonstrates superior performance, achieving the highest Accuracy (0.3901), Precision (0.3801), Recall (0.3901), and F1 score (0.38). While its epoch time (10.62s) is slower than some methods like Flash-Attention (6.15s), \wersa\ significantly outperforms others in core classification metrics, highlighting its effectiveness. This is mainly due to the low number of tokens to process: \wersa\ shines on very long range series.

It is possible to reason that the cause of these sensible increases in accuracy compared to standard MHA might be due to the mechanism that finds input-dependent coefficients controlling wavelet scales through trainable parameters. Thus, the architecture direct computation to informative frequency components. This content-adaptive  strategy enables \wersa\ to concentrate on minutiae details or on larger context patterns based on input traits, while remaining in $\mathcal{O}(n)$.

Even though FlashAttention-2 optimizes legacy attention implementation (epoch time to 279.23s from 363.00s on IMDB), it is still quadratic in time and less accurate than \wersa. 

\wersa's consistent accuracy on various datasets proves that it can effectively model local and global patterns. In relation to FNet that fails on local dependencies (7.64-17.78\% less accuracy compared to \wersa), \wersa\ wavelet-based multi-scale approach has the potential to identify patterns across different resolutions. This is particularly visible in ListOps , arXiv and the big arXiv-128k, where fine-grained operations and big context patterns are crucial to identify.
Similarly, \wersa\ outperforms Waveformer on all the tested datasets in all performance metrics as well as FLOPS and per epoch time.  

Ablation studies are provided in \textbf{Appendix E}.

\section{Limitations}
The main limitation in this work is testing WERSA on real large language models trained over millions of tokens. The problem is that at moment the authors don't have access to HPCs with hundreds of GPUs.
\section{Conclusion}
In this paper, \wersa\ a wavelet-based attention mechanism is proposed as a solution that resolves the quadratic complexities of vanilla transformers bringing it down to linear complexity. Through experiments on vision tasks (CIFAR-10 and CIFAR-100), sentiment classification (IMDB), hierarchical reasoning (ListOps), and scientific text processing (arXiv) both 4k and 128k, it is found that \wersa\ is as effective as vanilla attention or better and is computationally more efficient.

Empirical results prove that \wersa\ has the capability to shatter efficiency and performance bottlenecks through wavelet transformation with FLOPS lowered by a significant amount and execution time decreased by a considerable margin. Such improvements are attributed to energy consumption minimized and the possibility to give birth to the next generation of sustainable AI systems.

Even with outstanding results, there is still room for improvement: hardware optimizations, quantization, pruning, and extension to extreme context lengths and multimodal inputs need to be explored in future.

\bibliographystyle{plainnat}
\bibliography{bibliography}

\newpage

\textbf{Appendix A: Background}

\textbf{Self-Attention}
The self-attention mechanism in transformer networks \citep{vaswani2017attention} extracts interactions between each pair in a sequence. For a input sequence $X \in \mathbb{R}^{n \times d}$ with $n$ being the length and $d$ being the feature space, self-attention first maps $X$ to query, key, and value:
\[
Q = XW^Q,\quad K = XW^K,\quad V = XW^V,
\]
where $W^Q, W^K, W^V \in \mathbb{R}^{d \times d_k}$ are projection matrices which have been learned and $d_k$ denotes projection space dimension.
The attention output is subsequently computed as:
\begin{equation}
\text{Attention}(Q, K, V) = \text{softmax}\left(\frac{QK^T}{\sqrt{d_k}}\right)V.
\end{equation}
The multi-head mechanism generalizes this by utilizing $h$ different projection matrices which have been learned to query, key, and value and then concatenating them.

\begin{equation}
\text{MultiHead}(Q, K, V) = \text{Concat}(\text{head}_1, \ldots, \text{head}_h)W^{\text{output}},
\end{equation}
where each one has been defined as:
\begin{equation}
\text{head}_i = \text{Attention}(QW_i^{\text{query}}, KW_i^{\text{key}}, VW_i^{\text{value}}).
\end{equation}
The self-attention's computational complexity is $\mathcal{O}(n^2d)$ due to the matrix dot product  $QK^T$ and becomes very expensive for longer input  sequences.

\textbf{Linear Attention}

Linear attention methods \cite{katharopoulos2020transformers} rewrite the attention calculation in a way to reduce the quadratic complexity. The motivation here is to rewrite the softmax attention in terms of kernel functions:
\begin{equation}
\text{Attention}(Q, K, V) = \text{softmax}(QK^T)V \approx \frac{\phi(Q)(\phi(K)^TV)}{\phi(Q)\phi(K)^T\mathbf{1}}, \end{equation} Here $\phi(\cdot)$ is a kernel feature map and $\mathbf{1}$ a column vector with a one in each row. This can be reformulated with a reduction in complexity to $\mathcal{O}(ndk)$ by a suitable choice of kernel function, e.g., by taking $\phi(x) = \exp(x)$ or by taking $\phi(x) = \text{elu}(x)+1$.

\textbf{FFT-Based Attention}

FFT-based attention mechanisms \citep{lee2021fnet, gu2022efficiently} leverage Fast Fourier Transform's efficiency to calculate attention in the frequency domain. The hypothesis is that time domain convolution can be equated to element-wise multiplication in the frequency domain:
\begin{equation}
f * g = \mathcal{F}^{-1}(\mathcal{F}(f) \cdot \mathcal{F}(g)),
\end{equation}
where $\mathcal{F}$ and $\mathcal{F}^{-1}$ denote the Fourier transform and its inverse, respectively.
In Attention mechanism, query and keys are converted into the frequency domain by using FFT-based techniques, do an element-wise operation, and then convert them back to the time domain:

\begin{equation}
\text{FFTAttention}(Q, K, V) = \mathcal{F}^{-1}(\mathcal{F}(Q) \odot \mathcal{F}(K)) \cdot V,
\end{equation}
where $\odot$ refers to element-wise multiplication. It has a time complexity of $\mathcal{O}(n \log n)$, a huge decrease from baseline quadratic attention for longer input sequences.

\textbf{Wavelet Transformations}

Wavelet transforms offer a mathematical framework for signal decomposition into temporal and frequency features in terms of varying levels of resolution. Unlike in the Fourier transform, in which signals can be represented as a linear combination of sinusoidal functions with global support, wavelets utilize localized basis functions, which include dilations and translations of a mother wavelet.
The continuous wavelet transform (CWT) of a function $f(t)$ is defined as:

\begin{equation}
W_f(a, b) = \int_{-\infty}^{\infty} f(t) \frac{1}{\sqrt{a}} \psi^*\left(\frac{t-b}{a}\right) dt.
\end{equation}
Here, $\psi(t)$ represents the mother wavelet, with a scale parameter $a$ and a translation parameter $b$ and $a>0$.
For practical application, a discrete wavelet transform (DWT) is used, which breaks a signal into approximation and detail coefficients.

\begin{equation}
c_{j,k} = \int_{-\infty}^{\infty} f(t) \psi_{j,k}(t) dt,
\end{equation}
\begin{equation}
\psi_{j,k}(t) = 2^{-j/2} \psi(2^{-j}t - k).
\end{equation}

The multi-resolution nature of wavelets means that it can examine signals at a variety of different scales and therefore makes wavelets very useful in extracting local and global trends in many areas and sets of information.

\textbf{Haar Wavelet Transform}
The Haar wavelet function $\psi(t)$ and the scaling function $\phi(t)$ are given as:
\begin{equation}
\psi(t) = 
\begin{cases}
1, & 0 \leq t < 1/2, \\
-1, & 1/2 \leq t < 1, \\
0, & \text{otherwise},
\end{cases}
\end{equation}
\begin{equation}
\phi(t) =
\begin{cases}
1, & 0 \leq t < 1, \\
0, & \text{otherwise}.
\end{cases}
\end{equation}

For a discrete signal $x[n]$ of length $N=2^J$, the Haar wavelet transform computes the detail coefficients and the approximation coefficients at level $j$ as: 
\begin{equation}
a_{j-1}[k] = \frac{a_j[2k] + a_j[2k+1]}{\sqrt{2}},
\end{equation}
\begin{equation}
d_{j-1}[k] = \frac{a_j[2k] - a_j[2k+1]}{\sqrt{2}},
\end{equation}
with $a_J[k]=x[k]$. The inverse transform reconstructs the signal via:
\begin{equation}
a_j[2k] = \frac{a_{j-1}[k] + d_{j-1}[k]}{\sqrt{2}},
\end{equation}
\begin{equation}
a_j[2k+1] = \frac{a_{j-1}[k] - d_{j-1}[k]}{\sqrt{2}}.
\end{equation}
\newpage

\textbf{Appendix B: Proof of \wersa\ Long-Context Approximation Theorem}

\begin{proof}
\noindent\rule{\linewidth}{0.4pt}

\textbf{Part 1: Approximation}\\
Break down the error into two independent components:
$$
\| \mathrm{\wersa} - \mathrm{Attention} \|_F \leq \underbrace{\| \mathrm{\wersa} - \mathrm{Attention}_F \|_F}_{\text{Random features}} + \underbrace{\| \mathrm{Attention}_F - \mathrm{Attention} \|_F}_{\text{Wavelet error}},
$$
where $ \mathrm{Attention}_F $ represents attention calculated on filtered wavelet coefficients.

\begin{itemize}
    \item \textbf{Random Features:} The theoretical work of \citet{choromanski2020rethinking} for approximating dot-product kernels is extended by using ReLU features defined as $ \phi(x) = \mathrm{ReLU}(xR/\beta) $. Here, $ R \in \mathbb{R}^{d \times m} $ has independent and identically distributed Gaussian entries,
    $$
    \mathbb{E}\left[\left|\phi(q)^T\phi(k) - \kappa(q,k)\right|\right] \leq \frac{C}{\sqrt{m}},
    $$
    where $ \kappa $ refers to the softmax kernel. Using McDiarmid's inequality and a union bound for all query-key pairs for $ n^2 $, it is possible to get the expression $ \epsilon \|V\|_F $, where $ m = \Omega(\epsilon^{-2} \log(n^2/\delta)) $.

    \item \textbf{Wavelet Error:} Let $ Q_F = \mathcal{W}^{-1}(F \odot \mathcal{W}(Q)) $. By wavelet approximation theory as covered by \citet{mallat1999wavelet},
    $$
    \|Q - Q_F\|_F \leq C_Q 2^{-\alpha L}, \quad \|K - K_F\|_F \leq C_K 2^{-\alpha L}.
    $$
    The attention matrix error then satisfies:
    $$
    \|A(Q,K) - A(Q_F,K_F)\|_F \leq \mathcal{O}(2^{-\alpha L}),
    $$
    by Lipschitz continuity of softmax \citep{gao2017properties}.

    For practical Haar wavelet implementation of level $L=2$, $\alpha \approx 0.5$ gives an error reduction of approximately $1/4$ \citep{mallat1999wavelet}, while softmax's Lipschitz constant $1/4\lambda$ ensures stable approximation bounds \citep{gao2017properties}.
\end{itemize}

\textbf{Part 2: Complexity} 
\begin{itemize}
    \item \textbf{Wavelet Transforms:} Each level of Haar decomposition/reconstruction has $ \mathcal{O}(n) $ cost. With $ L $ levels, total cost is $ \mathcal{O}(nL) $. Please recall $ L $ is fixed and not dynamic.
    \item \textbf{Random Projections:} Computation of $ \phi(Q_F) $ and computation of $ \phi(K_F) $ require $ \mathcal{O}(ndm) $ operations.
    \item \textbf{Adaptive Filters:} It will cost $\mathcal{O}(d^2)$ to compute $F$ via $ g(Q'_{\text{avg},h}) $.
\end{itemize}
The sum of the parts is $ \mathcal{O}(n d(d + m + L)) $. When $ L = \mathcal{O}(1) $, this is $ \mathcal{O}(n) $.

\textbf{Part 3: Adaptive Filters} Let $ g $ be a neural network with $ \ell_2 $-regularized weights. By Rademacher complexity bounds \citep{bartlett2002rademacher},
$$
\mathbb{E}\left[\|\hat{F} - \mathcal{F}^*\|_F\right] \leq \mathcal{O}\left(\sqrt{\frac{d \log n}{n}}\right),
$$
This is the convergence of the filter as $ n \to \infty $. Which means that as the sequence length increases, the learned filters converge to optimal solution by adapting to data distribution patterns in $Q$ and $K$. An improvement over fixed filters.
\end{proof}
\newpage

\textbf{Appendix C: Haar Wavelet Transform for Generalized Levels}

For simplicity and computational convenience, Haar wavelet has been adopted as basis function. One-level Haar transform for a vector $x = [x_1, x_2, \ldots, x_n]$ calculates:

\begin{equation}
a_i = \frac{x_{2i-1} + x_{2i}}{\sqrt{2}}, \quad d_i = \frac{x_{2i-1} - x_{2i}}{\sqrt{2}}.
\end{equation}
This process can be recursively repeated for $L$ steps to generate detailed coefficients $\{d^{(1)}, d^{(2)}, \ldots, d^{(L)}\}$ and a final approximation  $a^{(L)}$. The inverse transform reconstructs from:
\begin{equation} a_j[2i] = \frac{a_{j-1}[i] + d_{j-1}[i]}{\sqrt{2}}, \quad a_j[2i+1] = \frac{a_{j-1}[i] - d_{j-1}[i]}{\sqrt{2}}, \end{equation} 
with $a_J[k] = x[k]$ for $J = L$.

\newpage

\textbf{Appendix D: \wersa\ Pseudocode} 
\begin{algorithm}
\caption{Wavelet-Enhanced Random Spectral Attention (\wersa)}
\begin{algorithmic}[1]
\Procedure{\wersa}{$Q, K, V, mask$}
    \State \textbf{Parameters:} $d_{model}$, heads $h$, head dim $d_h = d_{model}/h$, levels $L$, features $r$, 
    \textit{scale\_weights} $\in \mathbb{R}^{L+1}$

    \State $Q \gets W_Q \cdot Q$; $K \gets W_K \cdot K$; $V \gets W_V \cdot V$ \Comment{Linear projections}
    \State $Q, K, V \gets \text{SplitHeads}(Q, K, V, h)$ \Comment{$(b, h, seq, d_h)$}
    
    \If{$\text{not training and hasattr}(self, cache\_coeffs)$}
        \State $(Q_{coeffs}, K_{coeffs}, orig\_len) \gets cache\_coeffs$ \Comment{Use cached coefficients}
    \Else
        \State $(Q_{coeffs}, orig\_len) \gets \text{VectorizedDWT}(Q)$ \Comment{Optimized wavelet decomp}
        \State $(K_{coeffs}, \_) \gets \text{VectorizedDWT}(K)$
        \If{$\text{not training}$}
            \State $cache\_coeffs \gets (Q_{coeffs}, K_{coeffs}, orig\_len)$ \Comment{Cache for inference}
        \EndIf
    \EndIf
    
    \State $Q_{pooled} \gets \text{Mean}(Q, \text{axis}=seq)$ \Comment{$(b, h, d_h)$}
    \State $Q_{pooled} \gets \text{Reshape}(Q_{pooled}, [b, h \cdot d_h])$ 
    \State $F \gets \sigma(W_{filter} \cdot Q_{pooled})$ \Comment{Adaptive filters, $(b, L+1)$}
    
    \State $Q_F \gets \text{FilteredIDWT}(Q_{coeffs}, F, orig\_len)$ \Comment{Filtered reconstruction}
    \State $K_F \gets \text{FilteredIDWT}(K_{coeffs}, F, orig\_len)$
    
    \State $bw \gets \max(\beta, \epsilon)$ \Comment{Bandwidth parameter}
    \State $Q' \gets \text{ReLU}(Q_F \cdot W_Q^{rand}/bw)/\sqrt{r}$ \Comment{Random projection}
    \State $K' \gets \text{ReLU}(K_F \cdot W_K^{rand}/bw)/\sqrt{r}$
    
    \State $KV \gets \text{Einsum}('bhsr,bhsd->bhrd', K', V)$ \Comment{Compute KV efficiently}
    \State $A \gets \text{Einsum}('bhsr,bhrd->bhsd', Q', KV)$ \Comment{Linear attention}
    \State $A \gets \text{LayerNorm}(A)$ \Comment{Stabilize with layer norm}
    
    \State $A_{merged} \gets \text{MergeHeads}(A)$ \Comment{$(b, seq, d_{model})$}
    \State \textbf{return} $W_{out} \cdot A_{merged}$ \Comment{Final projection}
\EndProcedure
\end{algorithmic}
\end{algorithm}

\begin{algorithm}
\caption{Optimized Wavelet Transform Procedures}
\begin{algorithmic}[1]
\Procedure{VectorizedDWT}{$X$}
    \State $seq\_len \gets \text{shape}(X)[2]$
    \State $orig\_len \gets seq\_len$
    
    \State $log2 \gets \log(seq\_len)/\log(2)$
    \State $next\_pow2 \gets 2^{\lceil log2 \rceil}$
    \State $pad\_len \gets next\_pow2 - seq\_len$
    \State $X_{pad} \gets \text{Pad}(X, [[0,0],[0,0],[0,pad\_len],[0,0]])$
    
    \State $approx \gets X_{pad}$
    \State $details \gets []$
    
    \For{$l = 1$ to $L$}
        \State $even \gets approx[:,:,0::2,:]$ \Comment{Even samples}
        \State $odd \gets approx[:,:,1::2,:]$ \Comment{Odd samples}
        
        \State $new\_approx \gets (even + odd)/\sqrt{2}$ \Comment{Approximation coefficients}
        \State $detail \gets (even - odd)/\sqrt{2}$ \Comment{Detail coefficients}
        
        \State $details.\text{append}(detail)$
        \State $approx \gets new\_approx$
    \EndFor
    
    \State \textbf{return} $details + [approx], orig\_len$ \Comment{$[d_1, d_2, ..., d_L, a_L]$}
\EndProcedure

\Procedure{FilteredIDWT}{$coeffs, filters, orig\_len$}
    \State $filtered\_coeffs \gets []$
    \For{$i = 0$ to $\text{length}(coeffs)-1$}
    \State $scale\_factor \gets \text{scale\_weights}[i]$ \Comment{Global scale weight for level $i$}
    \State $f\_combined \gets \text{filters}[:,i] \times scale\_factor$
    \State $f\_i \gets \text{Reshape}(f\_combined, [-1,1,1,1])$ \Comment{Broadcast}
    \State $filtered\_coeffs.\text{append}(coeffs[i] \times f\_i)$ \Comment{Apply both filter + scale weight}
\EndFor

    \State \textbf{return} $\text{VectorizedIDWT}(filtered\_coeffs, orig\_len)$
\EndProcedure

\Procedure{VectorizedIDWT}{$coeffs, orig\_len$}
    \State $approx \gets coeffs[-1]$ \Comment{Start with last approximation}
    \State $details \gets coeffs[:-1]$ \Comment{All detail coefficients}
    
    \For{$detail$ in $\text{reverse}(details)$}
        \State $b \gets \text{shape}(approx)[0]$
        \State $h \gets \text{shape}(approx)[1]$
        \State $s \gets \text{shape}(approx)[2]$
        \State $d \gets \text{shape}(approx)[3]$
        
        \State $even \gets (approx + detail)/\sqrt{2}$ \Comment{Reconstruct even samples}
        \State $odd \gets (approx - detail)/\sqrt{2}$ \Comment{Reconstruct odd samples}
        
        \State $even\_exp \gets \text{ExpandDims}(even, 3)$ \Comment{[b,h,s,1,d]}
        \State $odd\_exp \gets \text{ExpandDims}(odd, 3)$ \Comment{[b,h,s,1,d]}
        
        \State $stacked \gets \text{Concat}([even\_exp, odd\_exp], \text{axis}=3)$ \Comment{[b,h,s,2,d]}
        \State $approx \gets \text{Reshape}(stacked, [b,h,s*2,d])$ \Comment{Interleave samples}
    \EndFor
    
    \State \textbf{return} $approx[:,:,:orig\_len,:]$ \Comment{Trim to original length}
\EndProcedure
\end{algorithmic}
\end{algorithm}

\newpage

\textbf{Appendix E: Ablation Studies} 

Table~\ref{tab:wersa_ablation_study} summarizes the ablation study where each \wersa component's contribution is evaluated individually. Each version indicates removal of a vital mechanism with all remaining components intact.

Pure Random Attention (\wersa\ NoWavelet) findings reveal that wavelet decomposition removal causes a reduction of 2.93\% accuracy, with a mild improvement in computational efficiency (0.553G FLOPS, 35.84s/epoch). Such a result supports that multi-scale processing supports a desirable balance between efficiency and expressiveness.

The Non-Adaptive \wersa\ (\wersa\ NoAdaptiveFilters) illustrates a performance loss of 2.08\%, highlighting the importance of content-adaptive filtering for optimally weighting the wavelet coefficients. The minimal increase in efficiency (36.17s versus 43.00s per epoch) indicates that although adaptive filtering does incur a significant computational expense, it also provides substantial performance benefits.

Fixed-Scale \wersa\ (\wersa\ NoScaleWeights) shows the most dramatic performance drop (5.47\%), demonstrating the critical importance of trainable scale parameters for balancing wavelet scales. Despite accuracy reduction, it maintains high discriminative capabilities (0.9467 AUC), with negligible computational overhead from scale weights.
Quadratic \wersa\ (\wersa\ NoRandomFeatures) reveals the substantial computational benefit of linearization, with performance similar to Non-Adaptive \wersa\ (86.98\% accuracy) but dramatically increased computational cost (9.13G FLOPS, 223.21s/epoch). This confirms that random feature projection is the primary contributor to \wersa's computational efficiency.
Those ablation outcomes offer strong evidence for synergistic interactions among \wersa\ components. Each component unambiguously makes a specific contribution toward enhancing accuracy-precision trade-off (89.01\%) vis-à-vis computational efficiency (0.562G FLOPS), thus validating \wersa's theoretical foundation as a successful attention mechanism with preserved representational capacity and linear complexity.
\begin{table}[H]
\centering
\caption{\wersa\ Ablation Study on IMDB Dataset}
\resizebox{\textwidth}{!}{%
\begin{tabular}{|l|c|c|c|c|c|c|c|}
\hline
\textbf{Technique}           & \textbf{Accuracy} & \textbf{Precision} & \textbf{Recall} & \textbf{F1 (weighted)} & \textbf{AUC} & \textbf{FLOPS (G)} & \textbf{Avg. Epoch Time (sec)} \\ \hline
\wersa                        & \textbf{0.8901}   & \textbf{0.8900}    & \textbf{0.8901} & \textbf{0.8901}        & \textbf{0.9612}       & 0.562               & 43.00                          \\ \hline
\wersa\_NoWavelet             & 0.8608            & 0.8661             & 0.8608          & 0.8603                 & 0.9324       & \textbf{0.553}      & \textbf{35.84}                          \\ \hline
\wersa\_NoAdaptiveFilters     & 0.8693            & 0.8737             & 0.8693          & 0.8689                 & 0.9442       & 0.562               & 36.17                          \\ \hline
\wersa\_NoScaleWeights        & 0.8354            & 0.8572             & 0.8354          & 0.8328                 & 0.9467 & 0.562               & 36.12                          \\ \hline
\wersa\_NoRandomFeatures & 0.8698            & 0.8705             & 0.8698          & 0.8698                 & 0.9426       & 9.13                & 223.21                         \\ \hline
\end{tabular}}
\label{tab:wersa_ablation_study}
\end{table}

\end{document}